\documentclass[9pt]{article}
\usepackage{spconf,amsmath,graphicx}

\usepackage{hyperref}
\usepackage{url}
\usepackage{xcolor}

\title{MUG: A General Meeting Understanding and Generation Benchmark}
\name{
\parbox{\linewidth}{\centering
Qinglin Zhang$^1$, Chong Deng$^1$, Jiaqing Liu$^1$, Hai Yu$^1$, Qian Chen$^1$\\
Wen Wang$^1$, Zhijie Yan$^1$, Jinglin Liu$^2$, Yi Ren$^2$, Zhou Zhao$^2$}
}
\address{$^1$ Speech Lab of DAMO Academy, Alibaba Group $^2$ Zhejiang University\\
  $^1${\tt \{qinglin.zql, dengchong.d, mingzhai.ljq, yuhai.yu\}@alibaba-inc.com} \\
  $^1${\tt \{tanqing.cq, w.wang, zhijie.yzj\}@alibaba-inc.com} \\
  $^2$ {\tt \{jinglinliu, rayeren, zhaozhou\}@zju.edu.cn}}
\usepackage[pscoord]{eso-pic}
\newcommand{\placetextbox}[3]{
\setbox0=\hbox{#3}% Put <stuff> in a box
\AddToShipoutPictureFG*{% Add <stuff> to current page foreground
\put(\LenToUnit{#1\paperwidth},\LenToUnit{#2\paperheight}){\vtop{{\null}\makebox[0pt][c]{#3}}}%
}%
}%

\begin{document}

\maketitle

\begin{abstract}
Listening to long video/audio recordings from video conferencing and online courses for acquiring information is extremely inefficient. Even after ASR systems transcribe recordings into long-form spoken language documents, reading ASR transcripts only partly speeds up seeking information. It has been observed that a range of NLP applications, such as keyphrase extraction, topic segmentation, and summarization, significantly improve users' efficiency in grasping important information. The meeting scenario is among the most valuable scenarios for deploying these \emph{spoken language processing} (SLP) capabilities.  However, the lack of large-scale public meeting datasets annotated for these SLP tasks severely hinders their advancement. To prompt SLP advancement, we establish a large-scale \textbf{general Meeting Understanding and Generation Benchmark (MUG)} to benchmark the performance of a wide range of SLP tasks, including topic segmentation, topic-level and session-level extractive summarization and topic title generation,  keyphrase extraction, and action item detection. To facilitate the MUG benchmark, we construct and release a large-scale meeting dataset for comprehensive long-form SLP development, \emph{the AliMeeting4MUG Corpus}, which consists of 654 recorded Mandarin meeting sessions with diverse topic coverage, with manual annotations for SLP tasks on manual transcripts of meeting recordings. To the best of our knowledge, the AliMeeting4MUG Corpus is so far the largest meeting corpus in scale and facilitates most SLP tasks.  In this paper, we provide a detailed introduction of this corpus, SLP tasks and evaluation methods, baseline systems and their performance\footnote{The Alimeeting4MUG corpus is released at \url{https://modelscope.cn/datasets/modelscope/Alimeeting4MUG/summary} and the baseline system is released at \url{https://github.com/alibaba-damo-academy/SpokenNLP}.}.
\end{abstract}

\placetextbox{0.5}{0.08}{\fbox{\parbox{\dimexpr\textwidth-2\fboxsep-2\fboxrule\relax}{\footnotesize \centering Accepted paper. \copyright  2023 IEEE. Personal use of this material is permitted. Permission from IEEE must be obtained for all other uses, in any current or future media, including reprinting/republishing this material for advertising or promotional purposes, creating new collective works, for resale or redistribution to servers or lists, or reuse of any copyrighted component of this work in other works.}}}

\section{Introduction}
\label{sec:intro}
Millions of meetings occur daily around the world, vital for seeking information, sharing knowledge, and improving productivity. They produce an enormous number of video/audio recordings. Listening to the long recordings for acquiring information is extremely inefficient. Even after ASR systems convert recordings into spoken language documents, reading these documents only partly ameliorates the efficiency issue. It has been observed that NLP technologies on spoken language documents, such as keyphrase extraction and summarization, are crucial for distilling, organizing, and prioritizing information and significantly improves users' efficiency in grasping important information, as shown in prior studies on topic segmentation~\cite{Ajideh06schema-theorybased}, keyphrase extraction~\cite{DBLP:journals/ir/Turney00} and summarization~\cite{DBLP:conf/coling/GurevychS04}. Topic segmentation and keyphrase extraction are also useful for downstream information retrieval~\cite{DBLP:conf/ht/SaltonSBM96,DBLP:conf/sigir/HearstP93,DBLP:journals/dss/GutwinPWNF99} and summarization~\cite{DBLP:conf/aaai/DiasAL07}.  

\begin{table*}[htb]
\begin{center}
\scalebox{0.95}{
\begin{tabular}{l l l l  l l l l}
\hline
Datasets    & \#Sessions & \#Avg. Turns   & \#Avg. Speakers   & Avg. Session Len. & Supported Tasks    & Language \\
\hline
AMI                             & 137       & 535.6     & 4.0           & 5,570.4   & Action, SUM, TS   & English \\
ICSI                            & 59        & 819.0     & 6.3           & 8,567.7   & Action, SUM, TS   & English \\
ELITR (English) & 120       & 727   & 5.9 & 7,066 & SUM & English \\
ELITR (Czech)   & 59        & 1,205     & 7.6   & 8,534 & SUM & Czech \\
QMSum       & 232       & 556.8     & 9.2           & 12,026.3  & QA, SUM, TS   & English \\ 
\hline
\textbf{AMC (ours)}  & \textbf{654}       & 376.3   & 2.5 & 10,772.5  & \textbf{Action, KPE, SUM,}   & Mandarin \\
& & & & & \textbf{Title, TS} & \\
\hline
\end{tabular}
}
\end{center}
\caption{\small{Statistics of existing meeting corpora and our \textbf{AliMeeting4MUG Corpus (AMC)}. For tasks supported by meeting corpora, \textbf{TS}:Topic Segmentation, \textbf{SUM}:Summarization, \textbf{QA}:Question Answering, \textbf{KPE}:Keyphrase Extraction, \textbf{Title}:Topic Title Generation, \textbf{Action}:Action Item Detection. Avg. Turns and Avg. Speakers are per session. Avg. Session length is the number of tokens per session after tokenization.}}
\label{tab:corpora}
\end{table*}

Compared to written\&formal text, meeting transcripts pose great challenges to spoken language processing (SLP) tasks. First, meeting transcripts exhibit \emph{a wide variety of spoken language phenomena}, such as disfluencies and redundancy, grammar errors,  coreference and information drop, colloquial expressions, and incomplete/fragmented sentences due to speaker interactions. These spoken language phenomena cause a significant discrepancy to the majority of training data of NLP models, i.e., written text, hence leading to dramatic performance degradation. Second, meeting transcripts are usually \emph{long-form documents} with several thousand words or more, challenging to mainstay Transformer-based models with quadratic complexity to the input sequence length. Although ASR errors also pose a great challenge to SLP on meetings~\cite{koay-etal-2020-domain}, we focus on the two key challenges above in this paper since many prior works demonstrate that even on manual transcripts of meetings, i.e., transcripts \textbf{without ASR errors}, performances of strong NLP models degrade dramatically compared to their performances on written text, due to these two key challenges. For example, abstractive summarization performance on Chinese written text dataset CLES~\cite{CLES} is about 20 absolute higher than that on our meeting corpus.

However, publicly available meeting corpora supporting SLP tasks are limited and on small scale, severely hindering the advancement of SLP in meetings. The ICSI meeting corpus~\cite{DBLP:conf/icassp/JaninBEEGMPPSSW03} comprises 75 recorded academic meetings of research discussions (~72 hours in total) at ICSI in Berkeley. The AMI meeting corpus~\cite{DBLP:conf/mlmi/CarlettaABFGHKKKKLLLMPRW05} consists of 137 meetings (100 hours in total) about industrial product design. The ELITR Minuting Corpus~\cite{nedoluzhko-etal-2022-elitr} consists of 179 project meetings, with 120 meetings in English and 59 in Czech. 
%Multiple minutes (bulleted lists of essential issues, actions, decisions, proposed activities) for each meeting are manually created. 
There are also publicly available committee meetings of formal discussions, which are quite different in domains and style from academic or project meeting corpora. The QMSum dataset~\cite{zhong2021qmsum} includes AMI and ICSI meetings as well as 25 committee meetings of the Welsh Parliament and 11 from the Parliament of Canada. Most other public meeting corpora are quite small and have no SLP annotations, such as the ISL meeting corpus~\cite{DBLP:conf/interspeech/BurgerMY02} and the NIST Meeting Room Corpus~\cite{DBLP:conf/mlmi/MichelAF06}.

To address this issue, we establish a \textbf{general Meeting Understanding and Generation benchmark (MUG)} for the community to benchmark the performance of a wide range of SLP tasks, including topic segmentation, topic-level and session-level extractive summarization, topic title generation,  keyphrase extraction, and action item detection. To facilitate the MUG benchmark, we construct and release a large-scale meeting dataset for long-form SLP development, \textbf{the AliMeeting4MUG Corpus (AMC)}, which consists of 654 recorded Mandarin meeting sessions with diverse topic coverage, with manual annotations for SLP tasks on manual transcripts of meeting recordings.  As compared to the existing meeting corpora in Table~\ref{tab:corpora}, to the best of our knowledge, AMC is so far the \textbf{largest meeting corpus in scale and facilitates the most SLP tasks}. We build baseline systems and report evaluation results on the MUG tasks. Next, we provide a detailed introduction to AMC, SLP tasks and evaluation methods, baseline systems and performance.

\section{The AliMeeting4MUG Corpus (AMC)}
\label{sec:corpus}
\subsection{Data Collection}
\label{subsec:collection}
\emph{The AliMeeting4MUG Corpus} extends the \textbf{224} meetings from our previously released \emph{Alimeeting} corpus for the ICASSP2022 M2MeT (multi-channel multi-party meeting transcription) challenge~\cite{DBLP:conf/icassp/YuZFXZDHGYMXB22} with \textbf{430} additionally collected meetings. Each meeting session consists of a 15-minute to 30-minute discussion by 2-4 participants covering certain topics.  \textit{We make sure all 654 meetings have no personally identifiable information, nor sensitive information about projects or organizations.} Topics span medical treatment, education, business, organization management, industrial production, and other daily routines. 
%Details of data collection for the \emph{Alimeeting} corpus are elaborated in~\cite{DBLP:conf/icassp/YuZFXZDHGYMXB22}. 
The extended meeting sessions generally adopt the same data collection procedure as the Alimeeting corpus~\cite{DBLP:conf/icassp/YuZFXZDHGYMXB22},  except that the extended meeting sessions have a smaller number of participants (from avg. 3.2 speakers per session in Alimeeting to 2.1 in the extended sessions) and a much lower speech overlap ratio (from 42.27\% and 34.76\% on Alimeeting Train and Eval sets to around 10\% in extended sessions). For creating AMC, we adopt the same manual transcripts with manually inserted punctuation and manual speaker labels for the Alimeeting corpus, and follow the same procedure for manual transcription, manual punctuation insertion and speaker labeling as conducted for the Alimeeting corpus~\cite{DBLP:conf/icassp/YuZFXZDHGYMXB22} on the extended sessions. We create manual annotations for all 5 SLP tasks (Table~\ref{tab:corpora}) for 524 meetings and manually annotate the rest 130 meetings with only topic segmentation.  For Track2-5 described in Section~\ref{sec:tracksetting-eval},  we partition the 524 meetings with all 5 SLP annotations into 295 sessions for training (\textbf{Train}), 65 sessions for system development (\textbf{Dev}), 82 sessions for Stage1 test set (\textbf{exceptTS-Test1}) and 82 sessions for Stage2(Final) test set(\textbf{exceptTS-Test2}). For Track1, the same Train and Dev sets are used and we partition the 130 meetings with only TS labels into 65 sessions for Stage1 test set(\textbf{TSonly-Test1}) and 65 sessions for Stage2(Final) test set(\textbf{TSonly-Test2}). Statistics of AMC Train and Dev sets are summarized in Table~\ref{tab:raw-data-stat}. 

\begin{table}[ht]
\begin{center}
\scalebox{0.8}{
\begin{tabular}{l l l}
\hline
 & \textbf{Train} & \textbf{Dev} \\
\hline
\#Sessions              & 295       & 65 \\
\hline
Avg. \#Speakers     & 2.7       & 2.6 \\
Avg. \#Turns            & 440.1     & 439.4 \\
Avg. \#Paragraphs       & 469.7     & 463.4 \\
Avg. \#Sentences        & 722.8     & 705.7 \\
Avg. Session Length     & 11253.5   & 11516.1 \\
Avg. Turn Length        & 25.6      & 26.2 \\
Avg. Paragraph Length   & 24.0      & 24.8 \\
Avg. Sentence Length    & 15.6      & 16.3 \\
\hline
\end{tabular}
}
\end{center}
\caption{\small{Statistics of Train and Dev sets of the AliMeeting4MUG Corpus. All average numbers are per-session average. All lengths are measured by the number of tokens after tokenization.}}
\label{tab:raw-data-stat}
\end{table}

\subsection{Spoken Language Processing Annotations}
\label{subsec:annotations}
All sessions in AMC are recorded with both far-field and near-field recordings. Before manual annotations, we align the signal from the near-field recordings with far-field recordings and select the signal with higher quality for manual transcription. 
%ASR output is used to assist transcribers to manually transcribe the audio using Praat\footnote{https://github.com/praat/praat}. 
We create manual transcripts for audio together with manually inserted punctuation and manual speaker labels, with careful quality control~\cite{DBLP:conf/icassp/YuZFXZDHGYMXB22}. Semantic units ended with a manually labeled period, question mark, and exclamation are treated as \textbf{sentences} for the following SLP annotations and baseline systems (Section~\ref{sec:baseline}). We use character tokenization for Chinese and whitespace tokenization for other languages.
Next, we segment the resulting spoken documents into paragraphs using our sequence model that sets new SOTA for paragraph segmentation on both written text benchmarks and spoken documents~\cite{DBLP:conf/asru/ZhangCLLW21}. The model improves performance on spoken document segmentation by exploiting phonetic information and improves efficiency by processing multiple sentences simultaneously and employing a self-adaptive sliding window. To simplify display, we break paragraphs on speaker changes based on the manual speaker labels. Then, we apply a series of manual annotations on the paragraph-segmented spoken documents, including Topic Segmentation, Topic-level and Session-level Extractive Summarization, Keyphrase Extraction, Topic Title Generation, and Action Item Detection. Our annotation guidelines fully consider synergy between tasks with details described as follows.

\noindent \textbf{Topic Segmentation (TS)} Following TS annotations on ICSI Meeting Corpus~\cite{DBLP:conf/sigdial/GruensteinNP05}, we consider purposes of TS as high-level understanding and presentation of discourse and enabling users to browse the content efficiently. Our topic segments denote a fine-grained, focused discussion of a subject matter.  The annotators first go through the entire document, acquiring a basic understanding of its content and discourse, then label key sentence candidates in the document for Topic-level Extractive Summarization. A valid topic segment is required to contain 1-3 key sentences conveying the central ideas of the topic.  Topic segments are non-overlapping, both semantically self-contained and discriminative from other topics. When some key sentence candidates among neighboring topics are semantically similar, these topics are scrutinized to determine whether they should be merged. Similar to~\cite{DBLP:conf/sigdial/GruensteinNP05}, we allow topics to cross speaker changes; whereas, in order to maintain the paragraph segmentation structures, a topic boundary is annotated only at the last sentence of paragraphs. TS-annotated documents are then annotated for Topic-level Extractive Summarization and Topic Title Generation \textbf{simultaneously}.

\noindent \textbf{Topic-level and Session-level Extractive Summarization (ES)} The core annotation schema for \emph{Topic-level Extractive Summarization} is described above. We restrict reference key sentences to sentence types other than interrogative sentences, and also conduct deduplication. To address incomplete meanings of key sentence candidates, their context sentences may be concatenated to create semantically complete key sentences. For \emph{Session-level Extractive Summarization}, the reference key sentences are based on deduplicating and filtering reference key sentences for Topics. Key sentence candidates that are not semantically complete are filtered out. We limit the number of session-level reference key sentences to be about \emph{twice} the number of annotated topics.

\noindent \textbf{Topic Title Generation (TTG)} Based on the annotations of TS and ES, annotators create a title for each topic by summarizing its central idea concisely with a phrase or short sentence, in formal language with an \textbf{(optional subject)-predicate-object} expression focusing on entities and events. TTG can be considered as \emph{Topic-level Abstractive Summarization into a very short summary}. TTG also serves quality assessment for ES: annotation errors on ES observed in this stage are fixed before annotating topic titles. A title may use phrases from annotated topic-level key sentences as well as contain information not covered by the key sentences, but may not create new information not conveyed in the topic. Also, titles for different topics should be significantly discriminative. 

\noindent \textbf{Keyphrase Extraction (KPE)}  Exploiting the annotations of TS, ES, and TTG, annotators extract top-K ($K<=20$) keyphrases (KPs) for a session. KPs need to be in formal language and semantically complete. The majority of KPs are entities and events.

\noindent \textbf{Action Item Detection (AID)} Following~\cite{DBLP:conf/sigdial/GruensteinNP05}, action items refer to a task discussed in the meeting and assigned to the participant(s) and expected to complete \emph{within a short time window} after the meeting. 
%Following major prior works, we formulate AID as a binary classification problem. 
The input to AID annotations is manual transcripts with manually inserted punctuation. We label sentences containing information about actionable tasks (task description, time frame, owner) as positive samples and otherwise negative samples.

\noindent \textbf{Exploring Multi-Annotator Annotations} For TS annotation, one annotator conducts initial annotations and an expert reviews and fixes the annotations to create the final manual topic labels. For each of ES, TTG, KPE, and AID annotations, each document is annotated by \textbf{three annotators}. We compute the standard ROUGE-1,2,L F-score ~\cite{lin-2004-rouge}(R-1,R-2,R-L)\footnote{We use https://pypi.org/project/rouge/ for $ROUGE-1,2,L$ computation.} to evaluate the Inter-Annotator Agreement (\textbf{IAA}) for ES and TTG between each pair of annotations. We compute exact F$_1$ (based on exact match) to evaluate IAA for KPE and compute the Kappa coefficient~\cite{carletta-1996-assessing} to evaluate IAA for AID.  For KPE, the union of the labels from the three annotators is used as the final manual labels for training and evaluation. For topic- and session-level ES,  the union of labeled key sentences from three annotators for each topic or session is used as a reference for training. For TTG, we create three copies of training data with reference from each annotator as a target and pool the data for training. For ES and TTG, we report average and best ROUGE scores based on the three versions of references. For AID, for inconsistent labels from annotators, an expert decides the final manual label for training and evaluation. Table~\ref{tab:annotation-stats} shows statistics and IAA of SLP annotations on AMC. The moderate IAA values indicate great challenges for SLP annotations on meetings, which demand more studies.

\begin{table*}[ht]
\begin{center}
%\scalebox{0.8}{
\begin{tabular}{c| c c |c| c| c| c | c}
\hline
& \multicolumn{2}{|c|}{\textbf{TS}} & Topic-level ES & Session-level ES & TTG & KPE & AID  \\
\hline
IAA     & \multicolumn{2}{|c|}{N/A}      & 49.53/30.50/41.13 & 55.65/28.40/34.97 & 30.79/16.63/28.17 & 55.62 & 0.50 \\
\hline
& \#Topics & Len. & Count/Topic & Count/Session & Len. & Count/Session & Count/Session  \\
mean    & 9.81 & 996.1    & 2.41  & 10.81 & 11.26 & 17.37 & 3.22  \\
std     & 2.22  & 353.9     & 0.66  & 2.93  & 1.85  & 3.53  & 3.86 \\
25\%    & 8     & 714     & 2     & 9    & 10    & 15    & 0 \\
50\%    & 9.5    & 950       & 3     & 10    & 11    & 17    & 2 \\
75\%    & 11    & 1230      & 3     & 12    & 13    & 20    & 5 \\
\hline
\end{tabular}
%}
\end{center}
\caption{\small{Statistics and Inter-Annotator Agreement (\textbf{IAA}) for all SLP annotations in the entire AliMeeting4MUG Corpus. Counts for KPE are computed on \textbf{union of three versions of annotations} while all other counts are based on individual annotations.
%Counts for Topic-/Session-level ES and KPE are computed on \textbf{union of three versions of annotations}. 
For evaluating IAA, we report ROUGE F-scores R-1/R-2/R-L for Topic-/Session-level Extractive Summarization (ES) and Topic Title Generation (TTG) tasks, exact F$_1$ for Keyphrase Extraction (KPE), and Kappa coefficient for Action Item Detection (AID).}}
\label{tab:annotation-stats}
\end{table*}
\vspace{-4mm}
\section{Track Setting and Evaluation}
\label{sec:tracksetting-eval}
In this section, we introduce the five tracks in the MUG benchmark for benchmarking the performance of five SLP tasks on AMC,  including task definitions and evaluation methods. 
%All tracks are constrained tracks, that is, for all the tracks, participants can only use publicly available pre-trained language models, the Alibaba Meeting corpus with all the SLP annotations provided by us, and publicly available corpora in our provided list for each track for self-supervised training or self-training. No extra publicly available data nor private data may be used for the challenge.

\noindent \textbf{Track 1: Topic Segmentation (TS)}
%\subsection{Track 1: Topic Segmentation (TS)}
%\label{subsec:topic-seg}
The Topic Segmentation (TS) track requires segmenting the manual transcripts of a session into a sequence of non-overlapping, topically coherent segments. For evaluation, we use three standard evaluation metrics, including positive F$_1$, $P_k$~\cite{DBLP:journals/ml/BeefermanBL99}, and \emph{WinDiff}(\emph{WD})~\cite{DBLP:journals/coling/PevznerH02}\footnote{We use https://segeval.readthedocs.io/en/latest/ for $P_k$ and \emph{WD} computation.}. Both $P_k$ and \emph{WD} use a fixed sliding window over the document, with the window size usually set to half of the average true segment length. $P_k$ is proposed to resolve the drawbacks of positive F$_1$, including the inherent trade-off between precision and recall and insensitivity to near-misses.  While sliding the window, $P_k$ determines whether the two ends of the window are in the same or different reference segments and increases a counter for a mismatch. The final score is computed by scaling the counter by the total number of measurements taken. However, $P_k$ still has problems, including sensitivity to variation in segment size distribution, penalizing false negatives more than false positives, and over-penalizing near-misses. To remedy these issues, \emph{WD} is proposed in~\cite{DBLP:journals/coling/PevznerH02} as an update to $P_k$. %\emph{WD} moves a fixed sliding window over the document, and penalizes the algorithm whenever the number of boundaries within the window does not match the true number of boundaries for that window of text. 
We use $1-P_k$ and \emph{1-WD} so that larger values indicate better performance, same as positive F$_1$.

\noindent \textbf{Track 2: Topic-level and Session-level Extractive Summarization (ES)}
%\subsection{Track 2: Topic-level and Session-level Extractive Summarization (ES)}
%\label{subsec:extractive-summarization}
The Topic-level and Session-level Extractive Summarization track requires extracting key sentences for each reference topic segment and the entire session, without modifying original sentences. We use the standard ROUGE-1,2,L F-scores~\cite{lin-2004-rouge}(R-1,R-2,R-L) as the evaluation metrics for the two subtasks of topic-level and session-level extractive summarization, respectively. Since the references are from three annotators, we report both average and best ROUGE-1,2,L scores based on the three references. We then average the ROUGE-1,2,L scores of the two subtasks as the final score for this track. Larger values indicate better performance.

\noindent \textbf{Track 3: Topic Title Generation (TTG)}
%\subsection{Track 3: Topic Title Generation (TTG)}
%\label{subsec:title-generation}
The Topic Title Generation track requires generating an informative and concise title for each reference topic segment. Similar to the ES track, we use the standard ROUGE-1,2,L F-score evaluation metrics and report both average and best ROUGE-1,2,L scores based on the three references.

\noindent \textbf{Track 4: Keyphrase Extraction (KPE)}
%\subsection{Track 4: Keyphrase Extraction (KPE)}
%\label{subsec:KPE}
The Keyphrase Extraction track requires extracting top-K keyphrases from a session that can reflect its main content. We compute \emph{exact} F$_1$ and \emph{partial} F$_1$ at top-K between the predicted KPs and the labeled KPs as evaluation metrics. Exact F$_1$ is computed when predicted KPs exactly match the labeled KPs. For computing partial F$_1$, we define two KPs $i,j$ as partially matched if $|LCS(i,j)|>=L$, where LCS denotes their longest common sequence and $L$ is set to 2 characters on this dataset. Partial F$_1$ is computed correspondingly based on the partial match.

\noindent \textbf{Track 5: Action Item Detection (AID)}
%\subsection{Track 5: Action Item Detection (AID)}
%\label{subsec:action-item-detection}
The Action Item Detection track requires detecting sentences containing information about actionable tasks as positive samples. We report positive precision, recall, and F$_1$ for this task.

\vspace{-4mm}
\section{Baseline Systems}
\label{sec:baseline}
\vspace{-2mm}
One key challenge of the MUG benchmark is handling long-form documents (Section~\ref{sec:intro}). The avg. session length of AMC is 10,772.5 tokens (Table~\ref{tab:raw-data-stat}). Hence, our baseline systems for all tracks carefully consider handling long-input text. Although Transformer~\cite{DBLP:conf/nips/VaswaniSPUJGKP17} has become the SOTA architecture for sequence modeling on a wide variety of NLP tasks and transformer-based pre-trained language models (PLMs) become dominant in NLP, such as BERT~\cite{DBLP:conf/naacl/DevlinCLT19}, GPT-2~\cite{radford2019language}, and RoBERTa~\cite{DBLP:journals/corr/abs-1907-11692}, the core self-attention mechanism has quadratic time and memory complexity to the input sequence length~\cite{DBLP:conf/nips/VaswaniSPUJGKP17}, limiting the max sequence length during pre-training (e.g., 512 for BERT) for a balance between performance and memory usage.  The standard approaches for BERT-series PLMs to handle long input are to truncate the input into the max sequence length, or to use a sliding window and aggregate the model results on each window as the final results. However, these approaches may degrade performance due to losing contextual information. 
Consequently, for \textbf{baseline systems for the three understanding tasks of Track 1/2/5}, to better handle long-form documents than BERT-series PLMs, we build a unified framework of fine-tuning a pre-trained efficient Transformer, Longformer-base~\cite{DBLP:journals/corr/abs-2004-05150}\footnote{https://huggingface.co/IDEA-CCNL/Erlangshen-Longformer-110M}.  Longformer provides a drop-in replacement of self-attention in Transformer, achieving linear complexity by using a window-based self-attention to capture local context and a global attention to encode inductive bias about the task. We model TS/ES/AID as sequence labeling tasks. 
%For Track 1/2/5 as sentence-level sequence labeling tasks, the global attention attends to the [CLS] token. 
We use a fixed sliding window with size 4096 and allow one sentence overlap. We set batch size as 2 and learning rate as 5e-5.  We also evaluate BERT-base\footnote{https://huggingface.co/bert-base-chinese} and Longformer-base with 512 sliding window size and find that Longformer-base with 4096 window size outperforms both of them on these tasks.
%and for question-answering tasks, the global attention attends to all question tokens. 
%In contrast, GA in PoNet is task-agnostic. 
%Instead, we investigate two efficient transformers, namely,  Longformer~\cite{DBLP:journals/corr/abs-2004-05150} and our previously proposed PoNet~\cite{DBLP:conf/iclr/TanCWZZL22}, both providing a drop-in replacement for self-attention in Transformer. To the best of our knowledge, our PoNet is the first work to explore the full potential of the simple pooling mechanism for token mixing and modeling long-range dependencies, by multi-granularity pooling, namely, global aggregation (GA), segment max-pooling (SMP), and local max-pooling (SMP), and pooling fusion to capture different levels of contextual information.  As shown in~\cite{DBLP:conf/iclr/TanCWZZL22}, our PoNet significantly outperforms Longformer on the long-range dependency modeling benchmark LRA, and demonstrates competitive transfer learning capabilities, reaching 96\% of BERT performance on GLUE. 
  %The computational complexity of Longformer is $O(N \cdot K \cdot d)$ and for PoNet it is $O(N \cdot d^2)$, where $N$, $K$, and $d$ denote the sequence length, the average number of tokens attended by each token, and the hidden dimension size, respectively.
\textbf{The baseline system for the generative TTG track} is based on fine-tuning the pre-trained BART-base model~\cite{DBLP:conf/acl/LewisLGGMLSZ20}\footnote{https://huggingface.co/fnlp/bart-base-chinese}. We truncate input to the max sequence length 512 of this model. We set learning rate as 3e-5 with warmup\_steps 300 and linear lr\_scheduler, batch size 16, and use label smoothing~\cite{DBLP:conf/cvpr/SzegedyVISW16} with $\beta=0.1$.  For baseline systems for TS/ES/TTG/AID, we use Adam optimizer and set dropout prob as 0.1. We train 10 epochs for TS/ES/AID and 5 epochs for TTG and select the best-performing model on the Dev set, respectively. \textbf{The baseline system for the KPE track} uses an unsupervised approach YAKE~\cite{campos2018yake}, which employs document-based statistical features for ranking candidates. Our earlier work~\cite{DBLP:conf/acl/ZhangCWDZLW022} shows that YAKE performs relatively stable on documents with varying lengths, especially on long documents, compared to many other unsupervised approaches. We evaluate YAKE\footnote{https://github.com/LIAAD/yake} with the default parameter settings (ngram=3, window\_size=1) except  Jieba~\footnote{https://github.com/fxsjy/jieba} for Chinese word segmentation on input documents (scoring exact/partial F$_1$ does not need to segment reference KPs) and an extended stopword list for both Chinese and English.
%with 2000 entries. 
Table~\ref{tab:baseline-results} reports baseline system results on the \textit{TSonly-Test1} set for Track1 and \textit{exceptTS-Test1} set for Track2-5 (Section~\ref{sec:corpus}) with evaluation metrics described in Section~\ref{sec:tracksetting-eval}. We find that TS performance on AMC \textbf{manual} transcripts is worse than that from applying the same baseline system on QMSUM \textbf{ASR} transcripts, and our baseline TTG results on AMC \textbf{manual} transcripts are worse than abstractive summarization SOTA on \textbf{ASR} transcripts of AMI,ICSI and QMSUM~\cite{DBLP:journals/corr/abs-2109-07943,DBLP:journals/corr/abs-2109-02492,DBLP:journals/corr/abs-2209-10052}, all suggesting that the SLP tasks on AMC are quite challenging.

\begin{table}[ht]
\begin{center}
\scalebox{0.7}{
\begin{tabular}{l l l l}
\hline
\multicolumn{4}{l}{\textbf{Track 1 Topic Segmentation (TS)}} \\
Model  & positive F$_1$     & $1-p_k$   & 1-WD \\
Longformer & $22.7_{\pm0.98}$    & $0.583_{\pm0.008}$    & $0.56_{\pm0.008}$ \\
\hline
\hline
\multicolumn{4}{l}{\textbf{Track 2 Extractive Summarization (ES) (AVG)}}  \\
Model  & R-1 Avg./Best    & R-2 Avg./Best   & R-L Avg./Best \\
Longformer & $53.83_{\pm0.39}$/$61.64_{_\pm0.68}$   & $32.33_{\pm0.60}$/$42.73_{_\pm0.84}$  & $42.94_{\pm0.61}$/$53.87_{_\pm0.68}$ \\
\hline
\multicolumn{4}{c}{\textbf{Topic-level ES}}  \\
Model  & R-1 Avg./Best    & R-2 Avg./Best   & R-L Avg./Best \\
Longformer &  $51.16_{\pm0.68}$/$63.0_{\pm1.03}$   & $34.4_{\pm0.78}$/$49.61_{\pm1.19}$   & $45.03_{\pm1.02}$/$59.61_{\pm1.2}$ \\
\hline
\multicolumn{4}{c}{\textbf{Session-level ES}}  \\
Model  & R-1 Avg./Best    & R-2 Avg./Best   & R-L Avg./Best \\
Longformer & $56.5_{\pm0.94}$/$60.28_{\pm1.2}$   & $30.26_{\pm0.77}$/$35.85_{\pm1.07}$   & $40.84_{\pm0.53}$/$48.13_{\pm0.43}$ \\
\hline
\hline
\multicolumn{4}{l}{\textbf{Track 3 Topic Title Generation (TTG)}}  \\
Model  & R-1 Avg./Best    & R-2 Avg./Best   & R-L Avg./Best \\
BART &  $32.16_{\pm0.21}$/$45.11_{\pm0.22}$   &  $17.87_{\pm0.22}$/$28.26_{\pm0.32}$  & $30.1_{\pm0.26}$/$43.16_{\pm0.22}$ \\
\hline
\hline
\multicolumn{4}{l}{\textbf{Track 4 Keyphrase Extraction (KPE)}}  \\
Model           & Exact/Partial F$_1$@10       & Exact/Partial F$_1$@15      & Exact/Partial F$_1$@20 \\
YAKE            & 15.2/24.9    & 17.5/27.8   & 19.1/29.5 \\
\hline
\hline
\multicolumn{4}{l}{\textbf{Track 5 Action Item Detection (AID)}}  \\
Model           &   positive P  & positive R    & positive F$_1$ \\
Longformer      &   $60.18_{\pm5.06}$           &   $66.89_{\pm3.29}$            &   $63.14_{\pm1.41}$                \\
\hline
\end{tabular}
}
\end{center}
\caption{\small{Baseline system performance on the \textit{\{TSonly,exceptTS\}-Test1} sets for the five tracks. Each mean and std are computed on results from 5 runs with different random seeds.}}
\label{tab:baseline-results}
\end{table}

\vspace{-7mm}
\section{Conclusion}
\label{sec:conclusion}
\vspace{-2mm}
We establish a general and comprehensive Meeting Understanding and Generation benchmark (\textbf{MUG}) to drive the spoken language processing (SLP) research on meetings. To facilitate MUG, we construct the AliMeeting4MUG Corpus. We define SLP tasks, build and evaluate baseline systems. Next, we plan to add tasks such as QA and Session-level Abstractive Summarization, cover more languages such as English, and facilitate multi-modality MUG research.
\vfil\pagebreak

\footnotesize
\bibliographystyle{IEEEbib}
\bibliography{strings,mybib}

\end{document}